\documentclass{article}

\usepackage{arxiv}

\usepackage[utf8]{inputenc} % allow utf-8 input
\usepackage[T1]{fontenc}    % use 8-bit T1 fonts
\usepackage{hyperref}       % hyperlinks
\usepackage{url}            % simple URL typesetting
\usepackage{booktabs}       % professional-quality tables
\usepackage{amsfonts}       % blackboard math symbols
\usepackage{nicefrac}       % compact symbols for 1/2, etc.
\usepackage{microtype}      % microtypography
\usepackage{array}
\usepackage{multirow}
\usepackage{enumitem}
\usepackage{xcolor}
\usepackage{float}
\usepackage{wrapfig}
\usepackage{amsmath}
\usepackage{amssymb}
\usepackage{graphicx}

\title{Deep and Dense Sarcasm Detection}

\author{
  Devin Pelser  \\
  The School of Mathematics, Statistics and Computer Science\\
  University of KwaZulu-Natal\\
  South Africa \\
  \texttt{215023955@stu.ukzn.ac.za} \\
   \And
 Hugh Murrell \\
  The School of Mathematics, Statistics and Computer Science\\
  University of KwaZulu-Natal\\
  South Africa \\
  \texttt{murrellh@ukzn.ac.za} \\
}

\date{}

\begin{document}
\maketitle

\begin{abstract}
Recent work in automated sarcasm detection has placed a heavy focus on context and meta-data. Whilst certain utterances indeed require background knowledge and commonsense reasoning, previous works have only explored shallow models for capturing the lexical, syntactic and semantic cues present within a text. In this paper, we propose a deep 56 layer network, implemented with dense connectivity to model the isolated utterance and extract richer features therein. We compare our approach against recent state-of-the-art architectures which make considerable use of extrinsic information, and demonstrate competitive results whilst using only the local features of the text. Further, we provide an analysis of the dependency of prior convolution outputs in generating the final feature maps. Finally a case study is presented, supporting that our approach accurately classifies additional uses of clear sarcasm, which a standard CNN misclassifies.
\end{abstract}

% keywords can be removed

\section{Introduction}

Sarcasm is a complex linguistic phenomenon in which the intended meaning of an utterance is not the same as its literal meaning \cite{karoui2017exploring}. Sarcasm's figurative and nuanced nature presents a challenging task within opinion mining and sentiment analysis \cite{Pang2007OpinionMA}. With the rise of social media use, and the need for understanding comments therein, affective computing has gained an increase in popularity. Existing systems for summarising reviews, or monitoring brand sentiment, often fail to detect the implicit meaning behind sarcastic remarks \cite{Davidov2010SemisupervisedRO} and consequently mislead the analysis of sarcastic phrases, with interpretations often being taken as literal. Consider the sentence ``\textit{Great! I love waking up sick!}''. We are easily able to recognise the sarcasm due to the presence of strong polarity shifts and common-sense; no one \textit{loves} waking up sick. The literal meaning of the sentence has been discounted, with the speaker expecting the listener to understand the implied intent. This, however, is not as easily identified within the context of machine learning due to the overall positive sentiment formed by the words `Great!' and `love'. Whilst certain sarcastic utterances are easily identified based solely on lexical and pragmatic cues dependencyring within, the need for extrinsic information has been identified in prior works \cite{Amir2016ModellingCW}. In the realms of social media, comments are often in response to a previous comment or in reference to a world event; suggesting context plays a key role in sarcasm detection. Further, the commonality of informal language and slang has been shown to diminish the reliance of grammatical hints \cite{Satapathy2017PhoneticBasedMN}. With past research finding difficulity in correctly classifying these particular phrases \cite{Poria2016ADL}.

The key aim of this work is to explore whether, perhaps, additional local cues could be extracted from the isolated utterance. In particular, we propose a deep neural network implemented with dense connectivity to build rich feature maps purely from the linguistic structure of the text. We do not seek to outperform the state-of-the-art, but rather, to determine if our model is able to rival past approaches, all of which make extensive use of both context and meta-data (such as user profiling to determine an author's sarcastic tendencies). This considerable usage of external information presents an issue for deploying real-world systems. Modelling the trends and opinions of each user, prior to categorising their comments, may not always be possible given privacy settings, or a lack of data \cite{Ghaeini2018AttentionalMS}. Additionally it introduces a large overhead, and places the focus on generating detailed profiles of users or forums, rather than the sarcastic characteristics of the utterance. Hence our work focuses solely on the language facet of sarcasm. Our approach uses multiple convolution layers - with direct connections between each - to facilitate the use of both low-level, simple features, as well as complex hierarchical ones, in order to develop a deeper understanding of the sarcastic text. We hypothesise that these diverse feature-maps would give further understanding of the sarcastic intent, or lack thereof, within a given phrase. Through empirical evaluations, we demonstrate our approach yields competitive results against existing models incorporating extrinsic material. We provide a demonstration of the model\footnote{https://nextjournal.com/Anon-Dem/dwenet-56}, as well as all datasets and source code using the NextJournal platform, readers may run the demonstration by creating an account and clicking `Remix'. In summary, the overall contributions of this work are as follows:

\begin{itemize}[noitemsep]
	\item Proposing a novel deep and dense sarcasm detection system to model the isolated utterance for classification.
	\item Examining the role of low level features in enhancing the accuracy of sarcasm classification.
	\item Presenting benchmark results on a new formal sarcasm dataset in the form of on ablation study.
\end{itemize}

\section{Related Work}

Whilst computational sarcasm is still a relatively new field - receiving an increased focus from researchers in recent times due to the massive growth of social media and the need for sentiment analysis therein - various approaches have been presented and examined. Tepperman et al. \cite{Tepperman2006yeahRS} studied detection within speech through the use of prosodic, spectral (average pitch of utterance, duration of utterance etc.) and contextual cues (gender, laughter etc.). Carvalho et al. \cite{Carvalho2009CluesFD} analysed comments on a Portuguese news site and found certain linguistic features were indicative of irony; such as emoticons, excessive punctuation and quotation marks. This statistical approach based on surface-level features has been further studied, with Gonz\'alez-Ib\`a\~nez et al. \cite{GonzlezIbez2011IdentifyingSI} exploring unigrams, dictionary-based lexical features and pragmatic features (punctuation, emojis etc.). Liebrecht et al. \cite{Liebrecht2013ThePS} extended the idea of n-grams as features through the use of unigrams, bigrams and trigrams, together with intensifiers to classify Dutch tweets. Given that a sarcastic utterance often implies the opposite to what is said, sentiment and semantic incongruity as a feature has also been investigated: Riloff et al. \cite{Riloff2013SarcasmAC} studied the presence of positive sentiment co-occurring with negative situational phrases and Buschmeier et al. \cite{Buschmeier2014AnIA} used the polarity between written words and the star rating on Amazon reviews. To build large self-annotated corpora Twitter API has been used to scrape comments containing tokens implying sarcastic intent (e.g. \#sarcasm) \cite{Reyes2013AMA}. This approach was followed by Pt\'acek et al. \cite{Ptcek2014SarcasmDO} with a feature set consisting of skipgrams and character n-grams to classify Czech and English tweets. The need for context was investigated by Wallace et al. \cite{Wallace2014HumansRC}, showing humans battle to infer ironic intent without it. The inclusion of extrinsic meta-knowledge has been increasingly used, with Bamman and Smith \cite{Bamman2015ContextualizedSD} making extensive use of contextual information regarding the author and audience. While Ghaeini et al. \cite{Ghaeini2018AttentionalMS} reported competitive results by incorporating context together with Bi-LSTMs and attention mechanisms. A manually annotated Reddit dataset was analysed by Wallace et al. \cite{Wallace2015SparseCI}, proposing features based on subreddit, sentiment, named entities and interactions between these. Automatic feature extraction through convolutions within NLP domains was first proposed by Collobert and Weston \cite{Collobert2008AUA}, demonstrating it's effectiveness across multiple tasks. Amir et al. \cite{Amir2016ModellingCW} extended this to sarcasm to mitigate the effort of handcrafting features; implementing a CNN to learn user embeddings based on an author's prior texts. Further deep learning methods have been reported with Poria et al. \cite{Poria2016ADL} yielding state-of-the-art results on a ensemble of an SVM preceded by four CNNs, three of which were pre-trained to extract emotion, sentiment and personality respectively. Combining a CNN, with an LSTM and DNN was also shown to improve performance against a recursive SVM \cite{Ghosh2016FrackingSU}. In recent times, Hazarika et al. \cite{Hazarika2018CASCADECS} extensively explored meta-data through profiling discussion forums based on all previously written comments, as well as capturing both stylometric and personality traits of the author through user embeddings. 

Within the field of computer vision, the application of residual learning \cite{He2015DeepRL} resulted in extremely deep networks which led to state-of-the-art results on multiple image datasets. This notion was further enhanced through dense connectivity by Huang et al. \cite{Huang2016DenselyCC}, showing that the dense variants outperformed the residual counterparts. Schwenk et al. \cite{Schwenk2016VeryDC} researched the application of deep residual archetypes to NLP domains, correlating an increase in depth with performance gain across numerous tasks such as; news categorisation, ontology classification and sentiment analysis. 

In this paper, deep densely connected networks for sarcasm classification are studied; noting that prior approaches make use of shallower models. Previous results indicate the necessity of contextual information and the insufficiency of purely textual cues. However, no contextual information is used in this work, with the primary research goal to ascertain whether our deeper network is able to extract richer sarcastic patterns intrinsic to the text. Furthermore, a formal dataset - free of noisy labels and colloquial language - is used to conduct analysis of the proposed approach compared to a CNN.

\section{Method}

Inspired by the promising results of deep residual based archetypes on multiple NLP domains \cite{Schwenk2016VeryDC}, together with the superior performance of dense connectivity within the field of computer vision; a deep, densely connected model is proposed for sarcasm classification. We hypothesise that the low-level features extracted through initial convolutions, in conjunction with abstracted hierarchical representations formed deeper in the network, will provide a richer understanding of the lexical, syntactic and semantic cues present within an isolated utterance. 
\begin{wrapfigure}[42]{r}{7cm}
	\centering
	\includegraphics[scale=1.05]{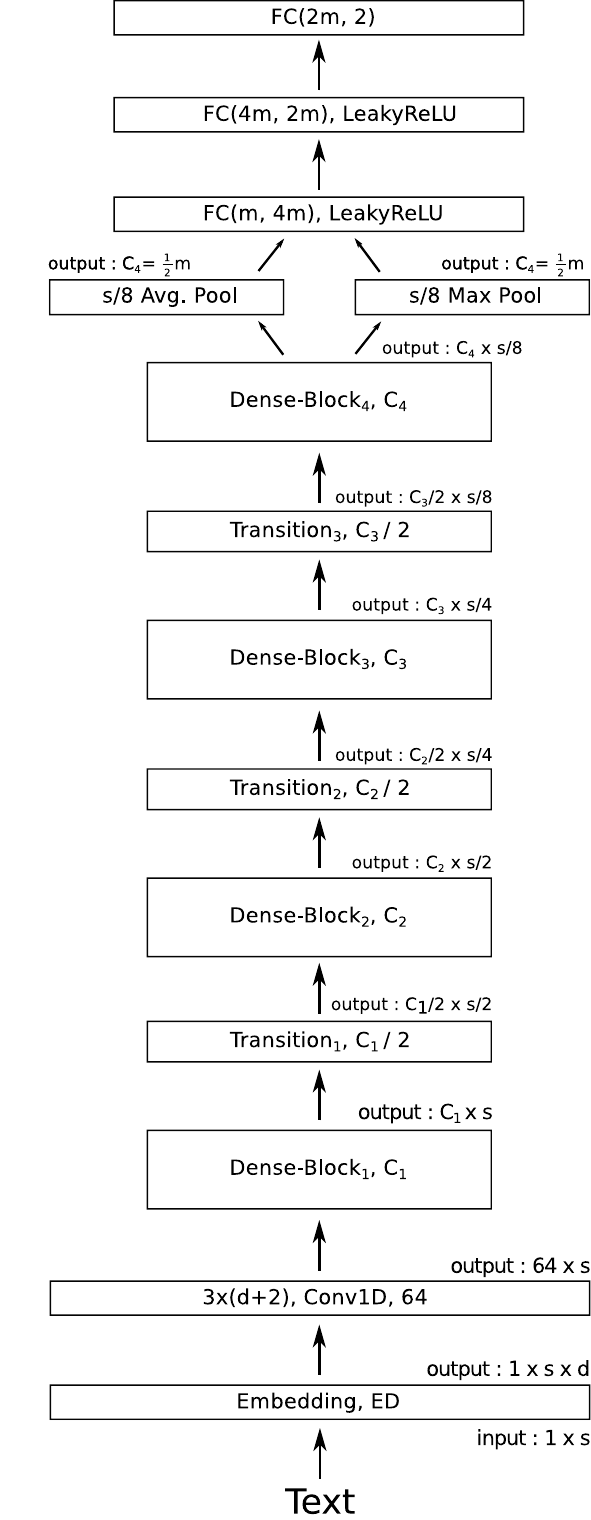}
	\caption{Overall network architecture.}\label{fig:prop_arch}
\end{wrapfigure}
Through empirical results we demonstrate that our deep model is capable of rivalling the performance of more complex networks which benefit from the use of contextual and extrinsic information. 

\subsection{Dense Connectivity}

The use of shortcut connections dependencyring between convolution layers was first proposed by Het et al \cite{He2015DeepRL}, facilitated by the element-wise addition of input and output tensors. While this was shown to be necessary in constructing very deep networks, the continual convolutions of low-level features resulted in distorted, high-level representations forming the final feature map. To allow the network to potentially use these low-level features, Huang et al. \cite{Huang2016DenselyCC} proposed the concatenation of prior convolution layer outputs to all sequential inputs; effectively allowing later layers to incorporate simple features in the production of their feature maps. We note that a convolution layer is defined as a Conv, BatchNorm, ReLU sequence.

Multiple densely connected convolution layers (dense-layers), with a filter size of 3, are stacked to form a dense-block. Each dense-layer within contributes $k$ channels to the overall block output, defined as the growth rate. The aforementioned concatenation consequently results in the input of consecutive dense-layers having $k$ additional channels. Hence, a dense-block of $n$ input channels, with $i$ dense-layers, will produce $n\,+\,(i\,\times\,k)$ outputs, noting that the signal dimension remains unchanged throughout. Downsampling is facilitated through the transition layers dependencyring between each block; halving the channel and signal dimension through convolutions of size 1 and average pooling with kernel size 2 respectively.

\subsection{Proposed Architecture}

The design of our network is shown in Fig.~\ref{fig:prop_arch}. The input text $W\,=\,[w_1,\,w_2,...,w_s]$ is initially passed to an embedding, responsible for generating vector representations, $\vec{u_j}\in \mathbb{R}^d$, for each word $w_j\in W$, where $d$ is the embedding dimension. Resulting in a single channel output tensor $[1,\,s,\,d]$, where $s$ is the number of tokens in input text $W$. An initial convolution layer follows, producing 64 channels by convolving along 3 embedding vectors at a time, effectively performing a 1-dimensional convolution through a 2-dimensional kernel which spans the entire embedding dimension (which has been padded once on both sides) - i.e a kernel of size $3\times(d+2)$. This yields an output tensor, $[64,\,s,\,1]$, formed from the 3-gram representations of the input text.
Four dense-blocks follow - outputting C$_i$ channels - with transition layers in between responsible for halving the signal and channel count. Consequently resulting in 4 hierarchical levels generating feature maps from complex combinations of the resulting signal, $s_i\,=\,\frac{s}{2^{i-1}}$, where $i$ is the $i^{th}$ dense-block. The final feature maps undergo a combination of both max and average pooling over the resultant signal. Ouputs are concatenated - yielding $m$ features - before being passing to three linear layers with leaky ReLU and dropout in between. Finally, classification dependencys through the softmax function at the head of the network.
The final model was implemented with a growth rate of 32 and dense-blocks of size 6, 12, 24 and 16 respectively.

\section{Experimental Results}

The proposed approach is tested on the binary classification of sarcastic text over several datasets. A formal dataset is used to conduct an analysis of the model compared to a CNN; in order to investigate the effect of dense connectivity and demonstrate the use of low-level features. Finally, the network is compared against several recent models, each incorporating contextual and meta-knowledge, on a dataset designed for context investigations. Thus, we demonstrate competitive results using only the lexical details within the isolated utterance.

\subsection{Datasets}

The Kaggle dataset \textit{Newsheadlines for Sarcasm Detection}\footnote{https://www.kaggle.com/rmisra/news-headlines-dataset-for-sarcasm-detection}, was constructed by collecting headlines from two news websites: TheOnion\footnote{https://www.theonion.com} and HuffPost\footnote{https://www.huffpost.com/}. The former is well known for producing satirical adaptations of world news and comprises the sarcastic entries, whilst the latter is a reputable news site and the headlines form the nonsarcastic entries. Consequently the resulting data is free of spelling mistakes, informal language usage and noisy labels; that is, the dataset can be considered as \textit{formal}. Furthermore, as opposed to datasets from Twitter and Reddit, the headlines are self-contained, they are not sent in response to a prior post or discussion. This allows us to focus entirely on the sarcastic characteristics without local context needing consideration. We analyse the performance of the proposed dense model in comparison to a baseline CNN to determine whether additional intrinsic features are indeed being detected. An outline of the dataset is seen in Table~\ref{table:datasummary}.

To evaluate the proposed approach against models incorporating extrinsic information, the self-annotated corpus for sarcasm - SARC\footnote{http://nlp.cs.princeton.edu/SARC/} (V2.0) - presented by Khodak et al. \cite{Khodak2017ALS}, is used. This dataset was designed for contextual investigations, with related works making considerable usage of said context. The dataset was constructed by scraping Reddit comments; with sarcastic entries being self-annotated by authors through the use of the $\backslash$s token, which indicates sarcastic intent on the website. Posts on Reddit are often in response to another comment; SARC incorporates this information through the addition of the parent comment and further child comments surrounding a post. Additional details about the author and which subreddit the post appeared on are also provided. We make use of only the original comments in the datasets, discarding the parent and additional child comments. Two variants of this dataset are considered for benchmarking: Main balanced and Political balanced - the latter consisting of comments obtained only from the political subreddit. Details of both datasets are presented in Table~\ref{table:datasummary}.

\begin{table}[h!]
	\centering
	\begin{tabular}{ m{4cm}  >{\centering\arraybackslash}m{2cm} >{\centering\arraybackslash}m{2cm}  >{\centering\arraybackslash}m{2cm}  >{\centering\arraybackslash}m{2cm}  }
		\multicolumn{1}{c}{}& \multicolumn{2}{c}{Training} & \multicolumn{2}{ c}{Testing} \\
		\cline{2-5}
		\multicolumn{1}{c}{}& \textit{non-sarc} & \textit{sarc} & \textit{non-sarc} & \textit{sarc}\\
		\hline
		Newsheadlines & 11988 & 9379 & 2997 & 2345\\ 
		
		SARC Main Balanced & 104209 & 109713 & 26173 & 27520\\ 
		
		SARC Pol Balanced & 6834 & 6834 & 1703 & 1703\\ 
		\hline
	\end{tabular}
	\caption{Dataset statistics for Headlines and SARC}
	\label{table:datasummary}
\end{table}

\subsection{Training and Testing Details}

Pre-trained GloVe\footnote{https://nlp.stanford.edu/projects/glove/} vectors of dimension 50 were found to be optimal and are used to initialise the non-static embedding layer. That is the word representations are updated during training to better align them for sarcasm detection. A batch size of 64 with a learning rate of 1e-03, together with the one cycle learning policy introduced by Smith \cite{Smith2015CyclicalLR} is adopted. A weight decay of 1e-02 is used throughout the investigations - with dropout at a rate of 0.2 added to the final 56 layer dense network for additional regularisation. Weights are intialised using the method outlined by He et al. \cite{He2015DelvingDI} and are updated through minimisation of the log-loss with the Adam optimiser \cite{Kingma2014AdamAM} having momentum range between 0.8 and 0.7. Furthermore, input texts are padded to a size of 64 for the Headlines dataset, whilst both Reddit datasets are padded to a size of 128 to facilitate the batched-learning within the convolutional layers. Texts larger than the indicated lengths are removed from the dataset. It must be noted that the padding token used has constant zero entries as its embedding and gradients are not tracked. All experiments are run 20 times with results presented as the average. The proposed model and investigations are implemented using Pytorch and the FastAI framework.

\subsection{Baselines}

Prior state-of-the-art approaches - for both Reddit datasets - which are compared against our work, are detailed below.

\begin{itemize}[noitemsep]
	\item \textbf{Bag-of-words:} An SVM network which receives the word-counts of the text as a vector of the size $V$, where $V$ is the vocabulary size.
	\item \textbf{CNN:} A simple CNN with 3 different filter sizes to extract n-gram features, as detailed by Hazarika et al. \cite{Hazarika2018CASCADECS}.
	\item \textbf{CNN-SVM:} An ensemble of 4 CNN's in which 3 are pretrained to extract sentiment, emotion and personality features from the given comment proposed by  Poria et al. \cite{Poria2016ADL}. The outputs are concatenated and passed to an SVM for the final classification. 
	\item \textbf{CUE-CNN:} Proposed by Amir et al. \cite{Amir2016ModellingCW}, user embeddings are modelled to obtain stylometric features which are then combined with a CNN.
	\item \textbf{CASCADE:} Stylometric and personality features of the authors are modelled and fused using canonical correlation analysis to obtain comprehensive user embeddings. Further, for each forum, all previous comments are combined to obtain discourse representations surrounding the topics and patterns therein. This approach was proposed by Hazarika et al. \cite{Hazarika2018CASCADECS} and holds the record results on both datasets.
	\item \textbf{CASCADE} (no personality features) \textbf{:} The above CASCADE model where user embeddings are generated without personality features.
	\item \textbf{AMR:} An RNN-based model incorporating BiLSTMs to model input comments and responses thereof. Attention mechanisms, projection and re-reading are also used to provide deeper representations \cite{Ghaeini2018AttentionalMS}.
\end{itemize}

\subsection{Results}

Here, we present the results of our model, dweNet, compared to that of other state-of-the-art approaches. Table~\ref{table:resultsummary} presents the yielded results on the two SARC datasets. We can see that all proposed approaches - including ours -  outperform the simple BOW and CNN baselines. Entries written in blue represent results which are rivaled by our network. Each model - except the full CASCADE implementation -  is challenged by our approach; this is interesting considering our method makes no use of context, nor meta-data. The full CASCADE model makes the most use of these extrinsic features; supporting the need for background information. However, the competitive results obtained by our model suggests that additional information is indeed available in the isolated utterance, allowing it to outperform a network which made use of pretrained models extracting emotion, sentiment and personality features. 

\begin{table}[h!]
	\centering
	\begin{tabular}{  m{6.5cm}  >{\centering\arraybackslash}m{1.6cm} >{\centering\arraybackslash}m{1.6cm}  >{\centering\arraybackslash}m{1.6cm}  >{\centering\arraybackslash}m{1.6cm}  }
		
		\multicolumn{1}{c}{}& \multicolumn{2}{c}{Main} & \multicolumn{2}{ c}{Pol} \\
		\cline{2-5}
		\multicolumn{1}{c}{}& \textit{Accuracy} & \textit{F1} & \textit{Accuracy} & \textit{F1}\\
		\hline
		Bag-of-words & \color{blue}0.63 &\color{blue} 0.64 & \color{blue}0.59 & \color{blue}0.60\\ 
		
		CNN & \color{blue}0.65 & \color{blue}0.66 &\color{blue} 0.62 & \color{blue}0.63\\ 
		
		CNN-SVM \cite{Poria2016ADL} & \color{blue}0.68 & \color{blue}0.68 & \color{blue}0.65 & \color{blue}0.67\\ 
		
		CUE-CNN \cite{Amir2016ModellingCW} & 0.70 & \color{blue}0.69 & \color{blue}0.69 & 0.70\\ 
		
		CASCADE \cite{Hazarika2018CASCADECS} & 0.77 & 0.77 & 0.74 & 0.75\\ 
		
		CASCADE (no personality features) & \color{blue}0.68 & \color{blue}0.66 & \color{blue}0.68 & 0.70\\ 
		
		AMR \cite{Ghaeini2018AttentionalMS} & \color{blue}0.68 & 0.70 & - & -\\ 
		\hline
		dweNet & 0.69 & 0.69 & 0.69 & 0.69\\
		\hline
	\end{tabular}
	\caption{Comparison of our approach, dweNet, with state-of-the-art models and simple baselines on two versions of the SARC dataset. Blue entries indicate results rivaled by our network and dashed lines where no results were reported.}
	\label{table:resultsummary}
\end{table}

\subsection{Ablation Study}

Experiments surrounding architectural designs are conducted on the proposed dense model to evaluate various features. Table~\ref{table:ablationsummary} details the results of all structural changes investigated on the Headlines dataset. Residual archetypes marginally outperformed the base CNN but resulted in lower accuracies and increased parameter counts when compared to the dense variant. This suggests low-level feature reuse is indeed significant, given that a ResNet is unable to take advantage of this. Further, depth was found to be beneficial in classification accuracy, resulting in an increase of 1.74\%. This, however, was expected based on promising depth investigations across several NLP tasks conducted by Schwenk et al. \cite{Schwenk2016VeryDC}; similar results were observed when increasing the growth rate. Both FastText\footnote{https://fasttext.cc/}\cite{mikolov2018advances} embeddings - standard and subword - were found to be suboptimal, with the 50D Glove representations increasing testing accuracy by over 2\%. Keeping the pretrained Glove embeddings static resulted in a notable drop of 2.99\%, which can be attributed to the nuanced nature of sarcasm. Pretrained embeddings are trained on large-scale typical language datasets, which will rarely include sarcastic phrases, resulting in word embeddings which do not align with the semantics of sarcasm.

\begin{table}[h!]
	\centering
	\begin{tabular}{  m{8cm}  >{\centering\arraybackslash}m{2cm} }
		
		\multicolumn{1}{c}{}& \multicolumn{1}{c}{Headlines} \\
		\cline{2-2}
		\multicolumn{1}{c}{}& \textit{Accuracy} \\
		\hline
		8 Layer CNN & 83.88  \\ 
		
		8 Layer ResNet & 84.21  \\ 
		
		8 Layer DenseNet & 85.55  \\ 
		\hline
		8 Layer DenseNet & 85.55  \\ 
		
		28 Layer DenseNet & 87.29  \\ 
		\hline
		8 Layer DenseNet GR = 4 & 85.55  \\ 
		
		8 Layer DenseNet GR = 32 & 86.85  \\ 
		\hline
		
		dweNet FastText-1M 300D & 86.51  \\ 
		
		dweNet FastText-1M-subword 300D & 86.15  \\ 
		\hline
		dweNet GLoVe 50 static & 85.68  \\ 
		\hline
		\hline
		dweNet GLoVe 50 non-static & 88.67 \\
		\hline
	\end{tabular}
	\caption{Comparison of our proposed approach, the final entry, to structural variants.}
	\label{table:ablationsummary}
\end{table}

\subsection{Low-Level Feature Use}

\begin{figure}[h!]
	\centerline{\includegraphics[scale=1]{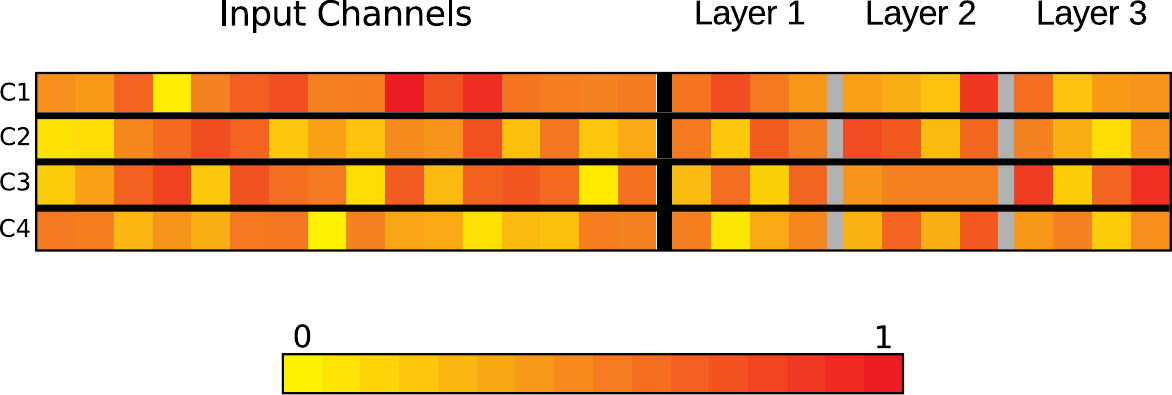}}
	\caption[Heatmap]{\label{fig:heatmap} 
		The average absolute filter weights of the convolution layers connected to the final layer in the fourth denseblock, channel by channel (C1 to C4). The colour encodes the normalised magnitude of these weights. The input feature maps are placed behind the vertical black bar, representing low-level features found earlier on in the network. Layer 1, 2 and 3 - separated by the grey rectangles - represent the initial three layers in the block.}
\end{figure}

A similar approach as Huang et al. \cite{Huang2016DenselyCC} is taken to support the notion that low-level features play a key role in increasing sarcasm detection when in combination with abstract hierarchical features. A 16 Layer DenseNet with block sizes (4, 4, 4, 4) and a growth rate of 4 is trained on the Headline dataset. The $L1$ norm of the weights connecting the preceding layers to the final layer in each block is taken in order to determine the dependency of the final layer with those prior to it. A detailed heatmap is shown in Fig~\ref{fig:heatmap} to visualise the intensity of this dependency for the final layer in the final block. From this we see the network assigns relatively strong importance to the input planes, indicating low-level features are indeed being used to form the final feature map. Furthermore the weights connecting the preceding 3 layers also exhibit high values, suggesting the abstract features also influence classification. The yellow blocks indicate features with little precedence and can be seen scattered throughout the heatmap; hinting that the features extracted by these layers are of little significance, or may be contained within another feature map.

\subsection{Case Study}

Results demonstrate that dense connectivity indeed allowed our model to rival related works which made extensive use of contextual information and meta-data, such as an author's post history and trends within a specific subforum. The 8 layer CNN was trained on the headlines dataset with incorrect classifications recorded. Similarly, we captured all errors made by our proposed model and the set difference was taken, that is, headlines which our model correctly identified, but the CNN did not. Analysis of these misclassifications was performed to determine whether our model was able to correctly classify additional headlines in which the sarcasm - or lack thereof - is clear based solely on the utterance. Below, we present a couple of these cases, paired with their actual label - sarcastic or nonsarcastic.

\begin{itemize}[noitemsep]
	\item \textit{Efforts of world's 16 billion chickens still not adding up to much.} - sarcastic
	\item \textit{CEO unveils bold new plan to undo damage from last year's bold new plan.} - sarcastic
	\item \textit{Like boxes of shit in your house? Get a cat.} - sarcastic
	\item \textit{United airlines temporarily suspends cargo travel for pets.} - nonsarcastic
	\item \textit{There have been more mass shootings this year than there have been days.} - nonsarcastic
\end{itemize}

The initial three headlines in the above list are all clearly sarcastic, with no background knowledge required to classify it as such. The latter two are seemingly normal news headlines with no sarcastic cues present. All five of which were incorrectly classified by the CNN, but identified correctly by our approach. This suggests that additional cues are available in the isolated text of the utterance than a standard CNN is able to extract. We do however note that our approach fails to classify text which require a clear understanding of the topic or background, such as: \begin{quote}\textit{Cops cleared on corruption charges after implicating decorated police dog.}\end{quote} This statement satirises cases where law enforcement were not held accountable for corruption, and further requires an understanding that a dog cannot be culpable of such an offense.

\section{Conclusion}

In this paper, we introduce a deep and dense network for extracting additional intrinsic information from a standalone utterance. Low-level features are shown to be used during the formation of the final feature maps. These, in combination with abstracted hierarchical features, enabled our model to rival state-of-the-art approaches which incorporated considerably more information on the SARC 2.0 datasets - such as user profiling and topic trends within a specific subforum. Our results demonstrate that whilst context is often needed to classify sarcasm; there is additional local information present that previous approaches have not taken advantage of.

\bibliographystyle{unsrt}  
%\bibliography{references}  

\end{document}